\documentclass[a4paper]{article}

\usepackage{fullpage} 
\usepackage{parskip} 
\usepackage{tikz} 
\usepackage{amsmath}
\usepackage{hyperref}
\usepackage{subfig}

\usepackage{graphicx}             
\usepackage{moreverb}             

\usepackage{graphicx}
\usepackage{amsmath,amssymb} 
\usepackage{color}
\usepackage{blindtext}
\usepackage{tabto, hyperref}
\usepackage{subfig, array}
\usepackage{blindtext, amsmath}
\usepackage{graphicx, tabto, hyperref}
\usepackage{subfig}
\usepackage{dblfloatfix}
\usepackage{graphics}
\usepackage{fontenc}
\usepackage{txfonts}
\usepackage{pifont}
\usepackage{color}
\usepackage{hyperref}
\usepackage{booktabs}
\usepackage{multirow}
\usepackage{float}
\usepackage{siunitx, array}
\usepackage{authblk}

\usepackage{float}
\usepackage{siunitx}
\usepackage[linesnumbered,ruled]{algorithm2e}

\usepackage{array}
\newcolumntype{P}[1]{>{\centering\arraybackslash}p{#1}}

\title{Distance Metric Learned Collaborative Representation Classifier}

\author[1]{Tapabrata Chakraborti}
\author[1]{Brendan McCane}
\author[1]{Steven Mills}
\author[2]{Umapada Pal}

\affil[1]{Dept. of Computer Science, University of Otago, NZ}
\affil[2]{CVPR Unit, Indian Statistical Institute, India}

\begin{document}

\maketitle

\begin{abstract}

Any generic deep machine learning algorithm is essentially a function fitting exercise, where the network tunes its weights and parameters to learn discriminatory features by minimizing some cost function. Though the network tries to learn the optimal feature space, it seldom tries to learn an optimal distance metric in the cost function, and hence misses out on an additional layer of abstraction. We present a simple effective way of achieving this by learning a generic Mahalanabis distance in a collaborative loss function in an end-to-end fashion with any standard convolutional network as the feature learner. The proposed method DML-CRC gives state-of-the-art performance on benchmark fine-grained classification datasets CUB Birds, Oxford Flowers and Oxford-IIIT Pets using the VGG-19 deep network. The method is network agnostic and can be used for any similar classification tasks.

\end{abstract}

\section{Introduction} 

Deep learning has achieved near human accuracy in recent years in many vision based tasks. However, any neural network inspired machine learning algorithm basically fits a function to given data using many parameters so as to learn discriminatory features from the input in an end-to-end manner. These features are then used to do the final discrimination operation using a standard distance metric. Though the network tries to learn the optimal feature space, it seldom tries to learn an optimal distance metric in the cost function, and hence misses out on an additional layer of abstraction [16,17].

The intuition for this work is that if the deep learned features are fed into a cost function with a distance metric which is also learned in tandem in an end-to-end manner, then it might help to further maximize the inter-class distance and help for such advanced classification tasks like fine-grained visual categorization. Deep convolutional networks are already proficient at recognizing base classes with sufficient data, but robust classification of sub-classes with fine-grained differences is still an open problem [1]. Thus as the representative problem to demonstrate our method, we choose fine-grained species recognition [2]. 

As the cost function, we use a collaborative representation classifier (CRC), which has recently been shown to be effective in fine-grained recognition [3]. CRC represents the test image as an optimal weighted average of training images across all classes and the predicted label is the class having least residual. CRC has a closed form solution; thus it is efficient and analytic. It is also a general feature representation-classification scheme and thus most popular features are compatible with it. 

The main contribution of this letter is to learn a generic distance metric in the cost function of a deep network in tandem with the learned features in an end-to-end manner. We provide an analytical derivation of the partial derivatives needed to optimise the distance metric and then back-propagate the gradients. The resulting system has wide generalisation since it is agnostic of the deep architecture and so can be used for any classification task. The method achieves state-of-the art results on three benchmark fine-grained species recognition datasets with the standard VGG-19 [9] deep network. We use standard publicly available models pre-trained on ImageNet [10] and fine-tuned on the three datasets, CUB Birds [5], Oxford Flowers [6] and Oxford-IIIT Pets [8], for fair comparison and ready reproducibility.

The rest of the paper is organized as follows. In Section 2, we present the original CRC in brief and the proposed DML-CRC in detail. Section 3 provides the experimental setup (the datasets, deep network and competing classifiers). Section 4 presents the experimental results with statistical analysis, followed by concluding remarks in Section 5.

\section{Proposed Method}
\label{sec:sec2}
In this Section, we present a brief description of the original formulation of the Collaborative Representation Classifiers (CRC) [4][20][21]. We then introduce in details the proposed Distance Metric Learned CRC (DML-CRC).

\subsection{Collaborative Representation Classifiers (CRC)}

Consider a training dataset with images in some feature space (such as one learned by a deep convolutional network) as $X=[X_1,\dots,X_c]\in \varmathbb{R}^{d \times N}$ where $N$ is the total number of samples over $c$ classes and $d$ is the feature dimension per sample. Thus $X_i \in \varmathbb{R}^{d \times n_i}$ is the feature space representation of class $i$ with $n_i$ samples such that $\sum_{i=1}^{c} n_i = N$.

The CRC model reconstructs a test image in the feature space $y \in \varmathbb{R}^d$ as an optimal collaboration of all training samples, while at the same time limiting the size of the reconstruction parameters, via the regularization term $\lambda$.

The CRC cost function is given as 
\begin{equation}
J(\alpha,\lambda)=\text{arg}\,\min\limits_{\alpha}\,(\|y-X\alpha\|_2^2+\lambda\|\alpha\|_2^2)
\end{equation}                  
where $\alpha=[\alpha_1,\dots,\alpha_c]\in\varmathbb{R}^N$ and $\alpha_i\in\varmathbb{R}^{n_i}$ is the reconstruction matrix corresponding to class $i$.

\noindent A least-squares derivation yields the optimal solution as
\begin{equation}
\hat{\alpha}=(X^TX+ \lambda I)^{-1}X^Ty
\end{equation}     

\noindent The representation residual of class $i$ for test sample $y$ can be calculated as: 
\begin{equation}
r_i(y)=\frac{\|y-X_i\hat{\alpha}_i\|_2^2}{\|\hat{\alpha}_i\|_2^2} \ \forall i \in {1,\dots,c}
\end{equation}

\noindent The final class of test sample $y$ is thus given by
\begin{equation}
C(y)=\text{arg}\,\min\limits_i\, r_i(y)
\end{equation}

\subsection{Distance Metric Learned CRC (DML-CRC)}

Most CRC methods, if not all, use the Eucledian $l_2$ norm or the Frobenius norm in the cost function. We replace it by a general Mahalanobis distance metric $\Sigma$ which can be optimised analytically, giving:

\begin{equation}
J(\alpha, \Sigma) = (y-X\alpha)^T\Sigma^{-1}(y-X\alpha) + \lambda\|\alpha\|_2^2 + \gamma\|\Sigma\|_2^2
\end{equation}


Let $X$ be the training set in some feature domain using the pre-trained deep model. Now, $y$ is each incoming image in the same feature domain, being used to fine-tune the network. Our aim is to find optimal $\Sigma$, $\alpha$ so as to minimize the cost function during the fine-tuning process. 

Differentiating $J$ with respect to $\alpha$, keeping $\Sigma$ constant we have:

\begin{equation}
\frac{\partial J}{\partial \alpha} = -2X^T\Sigma^{-1}(y-X \alpha) + 2 \lambda \alpha = 0
\end{equation}

Differentiating $J$ with respect to $\Sigma$, keeping $\alpha$ constant we have:

\begin{equation}
\frac{\partial J}{\partial \Sigma} = -\Sigma^{-1}(y-X \alpha)(y-X \alpha)^T \Sigma^{-1} + 2 \gamma \Sigma = 0
\end{equation}

Solving simultaneous equations 6 and 7, we have the new values of $\Sigma$ and $\alpha$ as:

\begin{equation}
\Sigma = \frac{\Sigma^{-1}(y-X \alpha)(y-X \alpha)^T \Sigma^{-1}}{2\gamma}
\end{equation}

\begin{equation}
\alpha = (X^T\Sigma^{-1}X + \lambda I)^{-1} X^T \Sigma^{-1} y
\end{equation}

During a specific round of back-propagation, once the new $\Sigma$ and $\alpha$ are set, the weights are then propagated back using the partial derivative with $X$ as follows.

\begin{equation}
\frac{\partial J}{\partial X} = -2\Sigma^{-1}(y-X \alpha)\alpha^{-1}
\end{equation}  

The DML-CRC training procedure for fine-tuning is presented in Algorithm 1. For further details on similar back-propagation schemes, the reader may refer to [18,19].

\begin{algorithm}[t!]
\SetAlgoLined
 \textbf{Initiate} reconstruction vector $\alpha$ and distance matrix $\Sigma$ \;
 \textbf{Extract} feature matrix $X$ using pre-trained model \;
 \For{each pass of fine-tuning}{
 \For{until termination condition reached}{
 Fix $\alpha$, update $\Sigma$ by eqn. 8 \;
 Fix $\Sigma$, update $\alpha$ by eqn. 9 \;
}
\textbf{Back-propagate} weights using eqn. 10 \;
  }

 \caption{DML-CRC fine-tuning algorithm}
\end{algorithm}

\section{Experimental Setup}
\label{sec:sec3}

In this Section, we describe the experimental setup: the datasets, chosen deep network and competing classifiers.

\subsection{Benchmark Datasets}

We have used three benchmark fine-grained species recognition datasets.

\textbullet \ \emph{CUB-2011 dataset:} It contains 11,788 images of 200 bird species [5]. The main challenge of this dataset is considerable variation and confounding features in background information compared to subtle inter-class differences in birds.

\textbullet \ \emph{Oxford Flowers dataset:} It has 8,189 images of 102 flowers, with at least 40 images per class [6]. It was developed by the Robotics Group at Oxford University. It is an expansion of the earlier dataset by the same group with 17 flower types with 80 images per class [7]. 

\textbullet \ \emph{Oxford Pets dataset:} This dataset, compiled by the Oxford Robotics Group and IIIT Hyderabad, consists of 37 categories of pet cats and dogs with around 200 images belonging to each class [8].

\subsection{Training on VGG-19 Deep Convolutional Network}

We have used the standard VGG-19 deep convolutional network from the Oxford Robotics group [9]. It has 19 layers, is trained on more than one million images from the ImageNet [10] dataset, and can classify up to 1000 object categories. We have fine-tuned the pre-trained VGG-19 model on our target datasets. For details of the training protocol, please directly refer to the benchmark work by Simon \emph{et al.} on neural constellation activations [11]. For fair comparison, we have used the baseline models provided by the authors of [11] in their GitHub repository [15]: pre-trained VGG-19 models on ImageNet and well as fine-tuned models on CUB Birds, Oxford Flowers and Oxford-IIIT Pets dataset using the CAFFE deep learning framework.

\subsection{Competing Classifiers}

We discuss two CRC based and two non-CRC based methods that have been used here for comparison. Note that all the methods have been used with VGG-19 features, but our method can be applied with any learned features.

\subsubsection{CRC based deep network classifiers}  There are many variants of CRC available; we choose patch based CRC (PCRC) as a major sub-class and probabilistic CRC (ProCRC) as a recent variant.


 \textbf{Patch based CRC (PCRC)} by Zhu \emph{et al}. [12] is a patch-based framework for collaborative representation (PCRC). Let the query image $y$ be divided into $q$ overlapping patches $y=\{y_1,\dots,y_q\}$. From the feature matrix $X$, a local feature matrix $M_j$ is extracted corresponding to location of patch $y_j$. Thus the modified cost function becomes: 

\begin{equation}
J(p_j,\lambda)=\|y_j-M_jp_j\|_2^2+\lambda\|p_j\|_2^2
\end{equation}  
where $M_j=[M_{j1},\dots,M_{jc}]$ are the local dictionaries for the $c$ classes and $\hat{p}_j=[\hat{p}_{j1},\dots,\hat{p}_{jc}]$ is the optimal reconstruction matrix for the patch $j$.

\textbf{Probabilistic CRC (ProCRC)} by Cai \emph{et al}. [13] is a probabilistic formulation of CRC where each of the terms are modeled by Gaussian distributions and the final cost function for ProCRC is formulated as maximisation of the joint probability of the test image belonging to each of the possible classes as independent events. The final classification is performed by checking which class has the maximum likelihood. 

\begin{equation}
J(\alpha,\lambda,\gamma)=\|y-X\alpha\|_2^2+\lambda\|\alpha\|_2^2+ \frac{\gamma}{K}\sum_{k=1}^{K}\|X\alpha-X_k\alpha_k\|_2^2
\end{equation}

\subsubsection{Non-CRC based classifiers used with VGG-Net} We use Constellation models due to its popularity in fine-grained recognition and also because we have used their pre-trained models directly for fair comparison. The other choice is the very recent paper on part attention models to compare against the state-of-the-art. 

\textbf{Constellation Neural Activations} by Simon \emph{et al.} [11] finds activation patterns with the help of convolutional networks in a completely unsupervised manner (no annotation or bounding box) to identify discriminatory parts for fine-grained classification. This is one of the popular baseline works in fine-grained classification and also provides the pre-trained models used in the current work.

\textbf{Object Part Attention Models} by Peng \emph{et al.} [14] is a very recently published work in fine-grained recognition and can be considered state-of-the-art. It reports results on the same datasets used in this work with VGG-19 features. This work combines an object level and a part level attention models with a spatial constraint that preserves spatial patterns.

\begin{table}[t!]

\renewcommand{\arraystretch}{2}

\caption{{{Classification results of proposed DML-CRC versus competitors on three fine-grained datasets (five-fold cross-validation mean percentage accuracy).}}}
\label{tab:tab1}
\centering
{
\begin{tabular}{|c|c|c|c|}

\toprule

      & \textbf{CUB Birds} & \textbf{Oxford Flowers} & \textbf{Oxford-IIIT Pets}  \\
      \midrule \hline

\textbf{CRC} & 75.24 & 91.83 & 83.30    \\
\hline
\textbf{PCRC} & 76.95 & 93.06 & 84.88    \\
\hline
\textbf{ProCRC} & 78.33 & 94.87 & 86.92    \\
\hline
\hline
\textbf{Constellation} & 81.01 & 95.34 & 91.60    \\
\hline
\textbf{OPAM} & 85.83 & 97.10 & 93.81    \\
\hline
\hline
\textbf{DML-CRC} & \textbf{88.49} & \textbf{98.65} & \textbf{95.12}    \\
\hline
\textbf{DML-ProCRC} & \textbf{89.95} & \textbf{99.33} & \textbf{96.58}    \\
\hline

\bottomrule

\end{tabular}
}
\end{table}

\begin{figure}[t!]
\centering

\subfloat[]{\includegraphics[width=1in,height=0.8in]{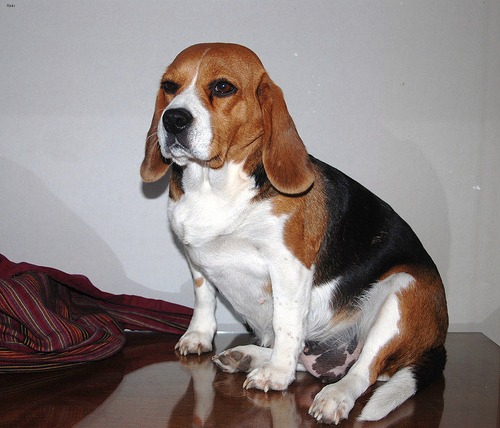}}
\hfil
\subfloat[]{\includegraphics[width=1in,height=0.8in]{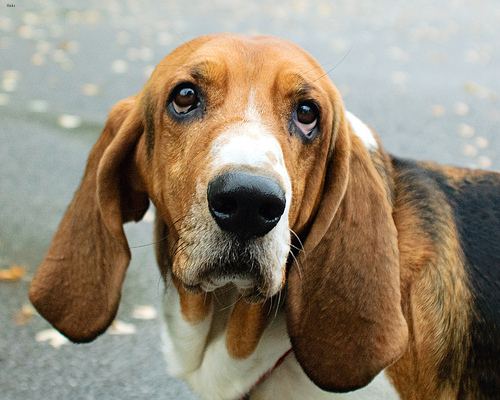}}
\hfil
\subfloat[]{\includegraphics[width=1in,height=0.8in]{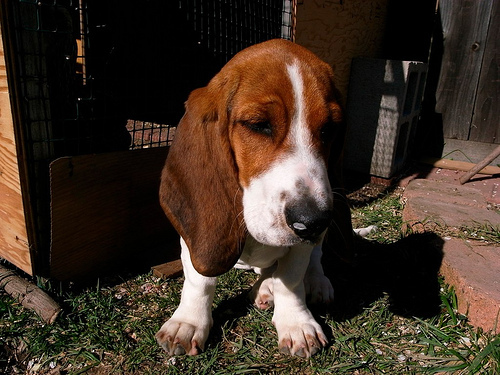}}
\hfill
\subfloat[]{\includegraphics[width=1in,height=0.8in]{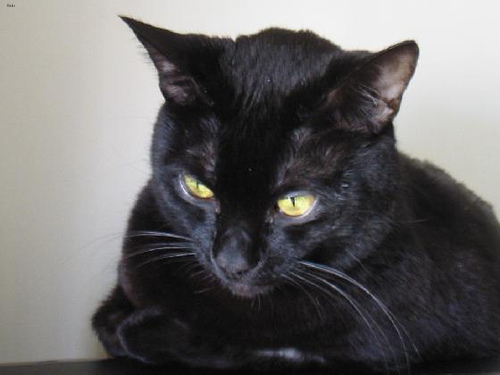}}
\hfil
\subfloat[]{\includegraphics[width=1in,height=0.8in]{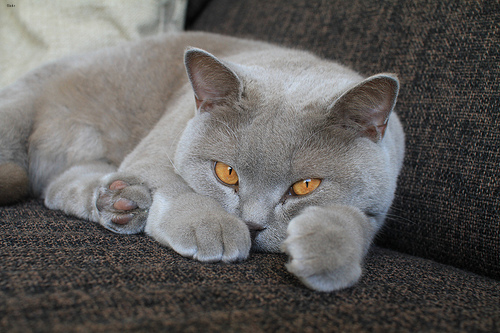}}
\hfil
\subfloat[]{\includegraphics[width=1in,height=0.8in]{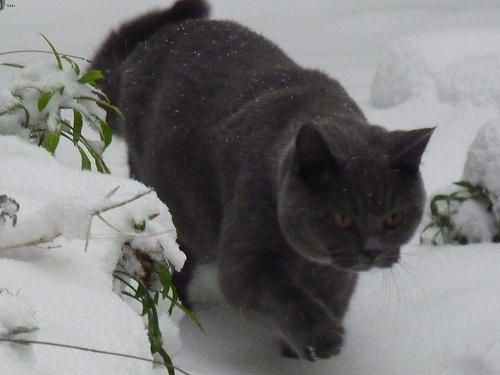}}

\caption{Qualitative results from the Oxford-IIIT Pets dataset. It has fine-grained image classes of cats and dogs. (a)-(c) give a misclassification example: (a) Basset Hound, (b) Beagle (characterised by longer ear) (c) Beagle, misclassified as Basset Hound by the proposed DML-CRC and its competitors, due to partial obfuscation of the discriminating longer ear of Beagle (both dogs have similar colored patchy skin). (d)-(f) give an example of correct classification: (d) Bombay Cat (e) British Shorthair (f) British shorthair correctly classified by proposed DML-CRC but misclassified by its competitors due to outlier black color of the cat (Bombay cat is generally black with narrower mouth while British Shorthair is gray with broader mouth).}
\label{fig:fig5}
\end{figure}

\section{Experimental Results}

For each dataset, experiments are conducted with five fold cross validation and percentage classification accuracies are presented in Table 1 with the accuracy of our method highlighted in bold. Among the CRC-based methods, basic CRC has the least accuracy and then there is an increase in the performance of the CRC variants. The proposed DML-CRC outperforms the original CRC and its variants comfortably. We also compare our method against two deep learning based methods, Constellation Model [11] and OPAM [14]. The rationale of choosing these two particular methods, have been discussed in previous Section. The proposed DML-CRC gives better results than both of these methods, thus establishing a new state-of-the-art. Fig. 1 presents qualitative results from the Oxford-IIIT Pets dataset.

It is important to note here that we have used the original CRC cost function first deliberately, to emphasize the contribution of the distance metric learning. This is demonstrated by the fact that even with vanilla CRC, we outperform the state-of-the-art albeit marginally in few cases. So it might be expected, that if a more recent version of CRC is used (like ProCRC), the margin of outperformance might increase. So we plug in the ProCRC cost function in place of the original CRC and the results are reported in Table 1, and as expected the performance improves further.

\section{Conclusion}
\label{sec:sec4}

We have shown that learning the distance metric for final discrimination of a convolutional network in an end-to-end manner enhances the performance of the system, keeping other factors like network architecture, data and training protocol constant. We used a collaborative representation based cost function (CRC) to evaluate an optimal Mahalanobis distance metric. A detailed analytic derivation is provided for partial derivatives of the CRC function. CRC has been recently shown to be effective for fine-grained classification and we achieve state-of-the-art results on three benchmark fine-grained classification datasets (CUB Birds, Oxford Flowers and Oxford-IIIT Pets). It should be further noted that the proposed method is agnostic of the deep architecture used and also may be utilised for any generic visual classification task. The current work uses vanilla CRC first deliberately for benchmarking; then we use the more recent ProCRC and observe improvement in performance. This suggests use of other new CRC formulations in future for possible further improvement in accuracy.

\end{document}